\documentclass{article}

\usepackage{arxiv}

\usepackage[utf8]{inputenc} 
\usepackage[T1]{fontenc}    
\usepackage{hyperref}       
\usepackage{url}            
\usepackage{booktabs}       
\usepackage{amsfonts}       
\usepackage{nicefrac}       
\usepackage{microtype}      
\usepackage{lipsum}		
\usepackage{graphicx}
\usepackage{natbib}
\usepackage{doi}

\title{Training of SSD(Single Shot Detector) for Facial Detection using Nvidia Jetson Nano}

\author{ {\includegraphics[scale=0.06]{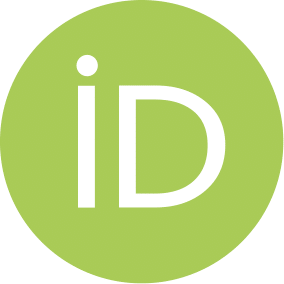}\hspace{1mm}Saif Ur Rehman}\\

	\texttt{saifurrehman4114@gmail.com} \\
	\And
	{\includegraphics[scale=0.06]{orcid.png}\hspace{1mm}Muhamamd Rashid Razzaq} \\

	\texttt{mrashidrazzaq143@gmail.com}\\
	
	\And
	{\includegraphics[scale=0.06]{orcid.png}\hspace{1mm} Muhammad Hadi Hussian} \\
	\texttt{hadihussain76@yahoo.com}
	
}

\hypersetup{
pdftitle={A template for the arxiv style},
pdfsubject={q-bio.NC, q-bio.QM},
pdfauthor={David S.~Hippocampus, Elias D.~Striatum},
pdfkeywords={First keyword, Second keyword, More},
}

\begin{document}
\maketitle

\begin{abstract}
In this project we have used computer vision algorithm SSD (Single Shot detector) computer vision algorithm and trained this algorithm from the dataset which consists of 139 Pictures. Images were labeled using Intel CVAT (Computer Vision Annotation Tool)

We trained this model for facial detection. We have deployed our trained model and software in NVIDIA Jetson Nano Developer kit. Model code is written in Pytorch deep learning framework. Programming language used is Python.	

\end{abstract}

\section{Introduction}
We are using NVIDIA Jetson Nano Developer kit  as our accelerator system.Which will contain Docker Container  which will contain the dataset and trained model SSD (Single Shot Detector) MobileNetV2 which we will be used to for facial detection.  

Video would be recorded through the Camera attached to the accelerator system. Code of the SSD (Single Shot Detector) MobileNetV2 is written in Python Programming Language and Deep learning framework which has been used is PyTorch.To optimized the neural network layers.NVIDIA TensorRT is used for faster Inference during the run time. NVIDIA TensorRT is built on the NVIDIA CUDA for parallel computing.

\section{Related Work}
Our Project is related to deep learning which have revolutionized the domain of AI.A subdomain of Machine Learning is deep learning. Deep learning means that you give data to the deep learning models to get the  features of the data to train the model which could then be used for inference for unseen data that's how the deep learning model is trained.

Deep learning field made it's when convolutional neural network \cite{ImageNet47:online} was trained to categorize the images in the ImageNet LSVRC-2010 contest. 

These days deep learning is being utilized everywhere from data science, big data, Image Classification, Image Segmentation, natural language processing, robotics, computer vision e.t.c

Deep learning models architectures have the input layer which is used to take the input data then the hidden layers which have perceptrons whose nodes are interconnected to each other to extract the features from the input data then the output layer comes in which is used to output the result  it is also called the inference layer.

In our project we are using SSD  deep learning model which is faster and accurate with smaller dataset then the RNN and FRCNN. When we compare the FRCNN \cite{1504080869:online} with  SSD \cite{1512023232:online} that SSD has a 76.9\%  mAP which was trained on PASCAL VOC cite{ThePasca65:online},COCO \cite{COCOComm41:online},and ILSVRC \cite{ILSVRC15} datasets individually to prove that the  FRCNN which is 66\%  mAP trained on PASCAL  is less accurate and faster.

\section{History Of Computer Vision}
Marvin Minsky made the first attempt to copy the human brain more than 50 years ago, making further research into the ability of computers to interpret knowledge to make wise decisions. The method of automating image processing has resulted in the programming of algorithms over the years. However, although there was acceleration in deep learning methods, it was only from 2010 onwards. Google Brain developed a neural network of 16,000 computer processors in 2012 that could recognize images of cats.

With the internet being a backbone, computer scientists have gained access to more knowledge than ever before. As costs of computer hardware continued to declined and improve. In the 1980s-90s, basic neural networks and algorithms emerged. The field of artificial intelligence, now more than half a century old, finally had its breakthrough moment at the.

The ILSVRC is an annual competition for image classification where research teams test their algorithms on the given data set, and then compete on multiple visual recognition tasks to achieve greater accuracy. A University of Toronto team then joined a deep neural network called AlexNet in 2012, making innovation for artificial intelligence and computer vision.

\section{Tensorflow and PyTorch}

Famous deep learning libraries are Tensorflow and PyTorch.Both Tensorflow and PyTorch are open-source. Tensorflow is primarily establish on Theano and has been originated from Google, while PyTorch is establish on Torch and has been originated from Facebook.

Critical distinction among them is the way they define the computational graphs. even as Tensorflow builds a static graph, PyTorch builds a dynamic graph.

PyTorch is more python favourite and AI models is easier to in it. However, for the usage of Tensorflow you may need to take training.

PyTorch changed in 2016 with the aid of facebook’s AI research lab. because it is normally intended for use in python, however it has a  C++ interface. TensorFlow supports many programming.  

TensorFlow uses TensorBoard for visualizing neural network.PyTorch on the other hand do not have a visualization feature.They uses Python packages for plotting.

PyTorch is the standard deep learning framework library for researchers, TensorFlow is preferred in commercial side. TensorFlow extensions for deployment on each servers and smartphone make this the desired alternative for teams that work with deep learning.

\section{Convolutional Neural Network}

An artificial neural network was inspired from the network of neurons present in the human brain. At each layer of artificial neural network input data is weights of previous artificial neural network is summed. Features of the input is taken out and prediction is made in the final layer.

\begin{figure}[!ht]
    \centering
    \includegraphics[width=1 
    \textwidth]{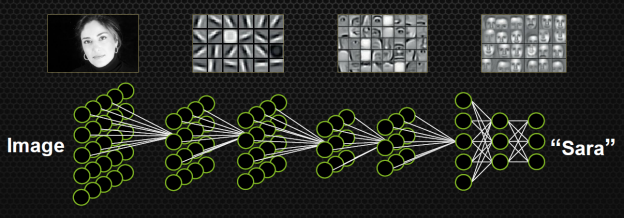}
    \caption{Processing of an Image in an Convolutional Neural Network \cite{artifici48:online}}
    \label{fig:artificial neural network}
\end{figure}

The convolution operation specific to CNNs combines the input data from one layer with a convolution filter to make a feature map for next layer. CNNs for image classification are generally composed of an input layer (the image), a series of hidden layers for feature extraction (the convolutions), and a fully connected output layer (the classification). Deep learning relies on Convolutional Neural Network (CNN) models to transform images into predicted classifications. A CNN is a class of artificial neural network that is made of  convolutional layers which extract the features from the input data, and is  preferred network for image applications

\begin{figure}[!ht]
    \centering
    \includegraphics[width=1 
    \textwidth]{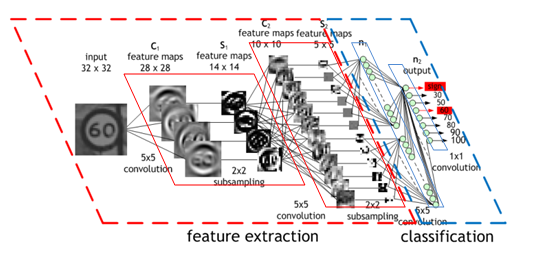}
    \caption{Convolutional Neural Network Architecture \cite{convolut0:online}}
    \label{fig:Convolutional Neural Network Architecture}
\end{figure}

As it is trained, the CNN adjusts automatically to find the most relevant features based on its classification requirements. 

\subsection{Single Shot Detector}

By using SSD \cite{1512023232:online}, we only uses one single shot to get objects in the image.On the other hand  regional proposal network (RPN) based models require two shots.

Implementing SSD in PyTorch for object detection, uses MobileNet backbones. SSD was released at 2016 and made improvements in object detection tasks with high accuracy, reaching over 74 mAP at 59 frames per second on datasets such as PascalVOC and COCO.

SSD do object detection and classification of it in the single forward pass.Detector is used to detect and distinguish the objetcs. Confidence Loss means how much sure network is sure about the detected object in the bounding box. Location Loss how much network prediction is away from training set prediction.

Features maps represents the major features in the Image. Multibox technique helps us to to detect the objects very clearly.

\begin{figure}[!ht]
    \centering
    \includegraphics[width=0.5 
    \textwidth]{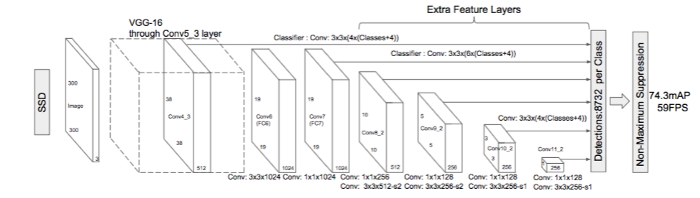}
    \caption{ SSD Architecture \cite{ssdarchi48:online}}
    \label{ssd_img}
\end{figure}

The authors of  SSD \cite{1512023232:online} argue that data augmentation is very important to improve the accuracy of the model.

\section{Jetson Nano Developer Kit}

Nvidia Jetson Nano Developer Board is very useful as it has GPU in it which helps us to run neural networks very fast for different applications in computer vision or robotics.

\begin{figure}[!ht]
    \centering
    \includegraphics[width=1 
    \textwidth]{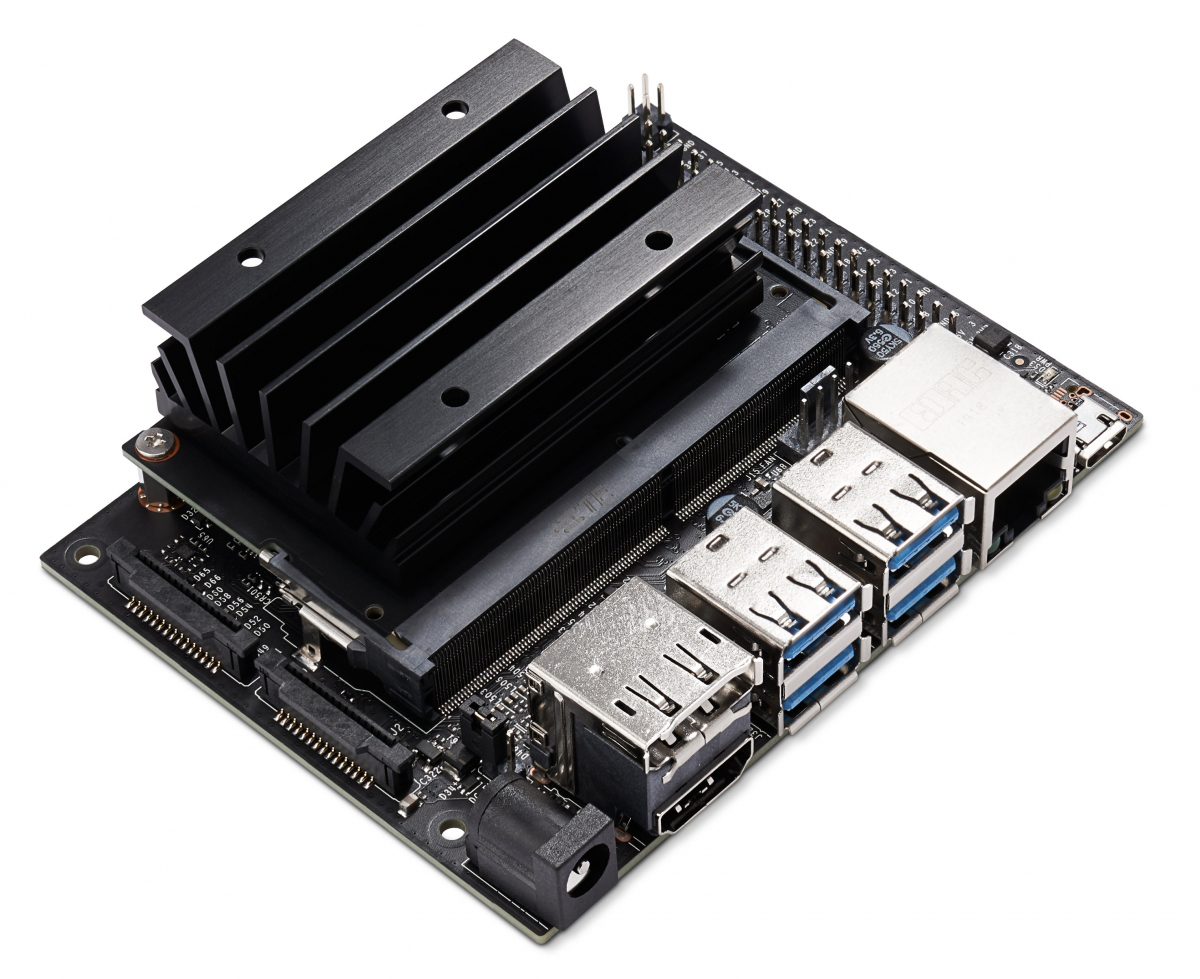}
    \caption{Jetson Nano Developer Kit \cite{JetsonNa51:online}}
    \label{fig:Jetson Nano Developer Kit}
\end{figure}

\begin{itemize}
    \item GPU	128-core Maxwell
    \item CPU Quad-core ARM A57 1.43 GHz
    \item Memory 4 GB 64-bit LPDDR4 25.6 GB/s
    \item  USB	4x USB 3.0, USB 2.0 Micro-B
    \item Display	HDMI and display port
    \item  Serial Communications Protocols GPIO, I2C, I2S, SPI, UART
    \item Board Power Rating 5V 2A, 5V 4A
    \item Storage 64GB  Solid State Drive (SSD)
    \item Operating System Ubuntu 18.04 
\end{itemize}

\section {A4Tech Webcam PK-810G}

Anti-glare coating to avoid reflections that are disturbing. Capture images of great quality even under low-light conditions. With no aliasing, intelligent multisampling provides fluent video transmission. Just plug it in and play, no installation of any software required. Intelligent MultiSampling. Snap Shot Buttom, free webcam driver, USB 2.0, built-in microphone.

\begin{figure}[!ht]
    \centering
    \includegraphics[width=.5 
    \textwidth]{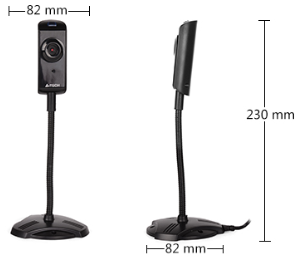}
    \caption{A4Tech Webcam PK-810G \cite{Antiglar30:online}}
    \label{fig:A4Tech Webcam PK-810G}
\end{figure}

\begin{itemize}
    \item Resolution: 480P, 640*480 Pixels
    \item Focus Range: 60cm and Beyond
    \item Built-in Mic.: 1 Mic.
    \item Output Video Format: MJPEG
    \item Frame Rate: 30fps

\end{itemize}

\section{CVAT}

CVAT is an open-source tool for annotating digital images and videos. The main function of the application is to provide users with suitable annotation tools. For that purpose, we designed CVAT as a handy service that has many great features.
CVAT is a browser-based application for both individuals and teams that supports different work scenarios. The main tasks of supervised machine learning can be divided into three groups:
\begin{itemize}
    \item Object detection 
    \item Image classification

    \item Image segmentation
\end{itemize}

CVAT allows you to annotate image/video for each of these cases. Here are some advantages and disadvantages of the tool.

\subsection{Advantages}
    \begin{itemize}
        \item Web-based. In this technique Users don’t need to install the CVAT app; User just have to run the tool link in a browser if user want to create a task or annotate data.
\item Collaborative. They can create a public project and split the project work between other users.
\item Easy to deploy. CVAT can be deployed in the local network using Docker.
\item Deep Learning Deployment Toolkit (Intel® Distribution of OpenVINO™ toolkit element) 
    \end{itemize}
\subsection{Disadvantages}
    \begin{itemize}
        \item Limited browser support. CVAT’s users works only on Google Chrome Platform. CVAT is not working in other browsers, but it may work on Chromium based browsers like a Opera Browser.
        \item All test have to be done manually, considerably slowing the development process.
    \end{itemize}
    
\section{Our Approach}

We trained this computer vision algorithm and deployed in the Jetson Nano.Training took around 2 hours. We were using it full performance.

Hyperparameter were that we set the epoch at 5 and the iteration was of 2 in each epoch.There was 139 total data set,validation dataset was around 29, and the training data set was around 110.

Learning rate was of 0.01 of base net layer and of extra layer. There was 3 classes for facial detection Saif, Hadi, and Rashid. Background is always added by deafult.

Accuracy when tested was approximately 97

\begin{figure}[!ht]
    \centering
    \includegraphics[width=1 
    \textwidth]{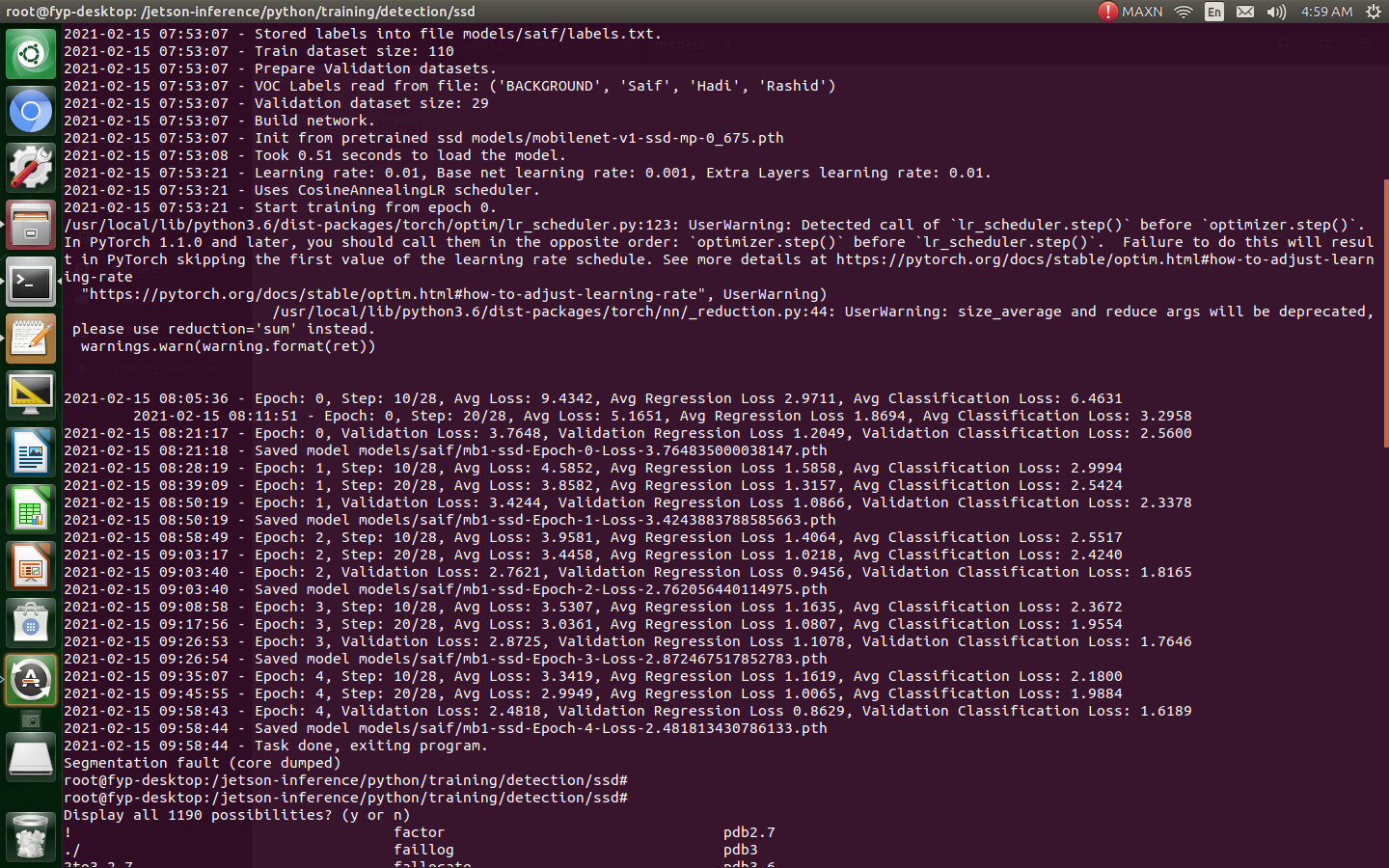}
    \caption{Training of SSD Model}
    \label{fig:SSD Training on Jetson Nano}
\end{figure}

\begin{figure}[!ht]
    \centering
    \includegraphics[width=0.5 
    \textwidth]{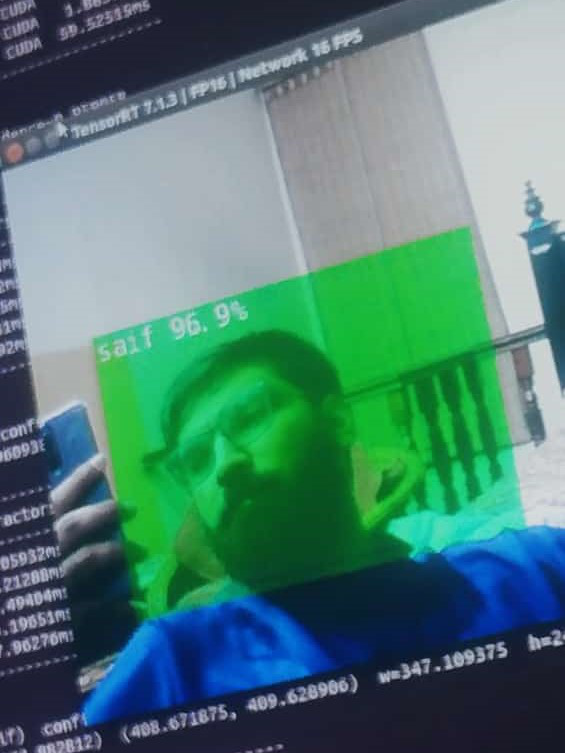}
    \caption{Facial Detection}
    \label{fig:Detection of Facial Detection}
\end{figure}

\vfill

\section{Conclusion and Future Work}

You can see that for facial detection SSD(Single Shot Detector) algorithm is really a great algorithm for facial detection.It's accuracy with small dataset and training very quickly makes it's a very effective algorithm.

Our future work would be used it in a embedded system and utilize it for real world application and solve a real world problem.
\bibliographystyle{unsrtnat}
\bibliography{references}  






\end{document}